# Assessing the Quality of Multiple-Choice Questions Using GPT-4 and Rule-Based Methods

Steven Moore[1][0000-0002-5256-0339], Huy A. Nguyen[1][0000-0002-1227-6173], Tianying Chen[0009-0003-9321-4205], John Stamper[0000-0002-2291-1468]

[1] Carnegie Mellon University, Pittsburgh PA 15213, USA
StevenJamesMoore@gmail.com

**Abstract.** Multiple-choice questions with item-writing flaws can negatively impact student learning and skew analytics. These flaws are often present in student-generated questions, making it difficult to assess their quality and suitability for classroom usage. Existing methods for evaluating multiple-choice questions often focus on machine readability metrics, without considering their intended use within course materials and their pedagogical implications. In this study, we compared the performance of a rule-based method we developed to a machine-learning based method utilizing GPT-4 for the task of automatically assessing multiple-choice questions based on 19 common item-writing flaws. By analyzing 200 student-generated questions from four different subject areas, we found that the rule-based method correctly detected 91% of the flaws identified by human annotators, as compared to 79% by GPT-4. We demonstrated the effectiveness of the two methods in identifying common item-writing flaws present in the student-generated questions across different subject areas. The rule-based method can accurately and efficiently evaluate multiple-choice questions from multiple domains, outperforming GPT-4 and going beyond existing metrics that do not account for the educational use of such questions. Finally, we discuss the potential for using these automated methods to improve the quality of questions based on the identified flaws.

## 1 Introduction

Multiple-choice questions (MCQs) are a widely used form of assessment in higher education, both for formative and summative evaluations. MCQs are advantageous because of their efficiency to score, objective grading, ability to generate item-analysis data, and the shorter time required for students to respond [5]. In recent years, the task of authoring educational MCQs is no longer specific to instructors, and the popularization of automatic question generation (AQG) systems further scaled up this process [21]. Another method for scaling the creation of educational MCQs is having students take part in the process of question creation, commonly referred to as a form of learnersourcing [35]. Student-generated questions often have higher quality, and target more complex cognitive processes compared to AQG [14]. The process of generating questions also has educational benefits for students and can lead to positive learning outcomes, such as improved retention and transfer [19].

While student-generated questions typically have higher quality than those created through automated methods, their quality may widely vary due to multiple uncontrollable factors [28]. On one hand, poorly designed MCQs may exhibit characteristics that can be exploited by pattern recognition and guessing, thus leading to shallow learning [9]. On the other hand, ensuring high-quality MCQs, whether created by AQG or students, is itself a challenging task. Common evaluation methods used by previous research include using experts, other students, or automated methods [25]. Even though automatic methods are most efficient, they come with important caveats. Notably, existing automated methods often rely on the surface-level features of a question, such as the readability of text length, without considering the pedagogical value it adds to the assessment [2]. Additionally, these methods are often applied to datasets consisting of questions targeted at lower academic grade levels, such as basic reading comprehension, or questions that are not used in an educational context at all [14]. While the use of human experts might provide the most accurate assessment of question quality, the manual evaluation process often lacks standardization and efficiency [21]. However, human evaluation can provide a more in-depth analysis, considering the question's potential to support learning. For instance, the Item-Writing Flaws (IWFs) rubric is an effective evaluation method which considers the pedagogical value of the question and its answer choices through various criteria [3, 8]. This rubric typically consists of 19 items that assess whether an educational MCQ is acceptable for use in the classroom or not. However, applying this rubric to a large number of questions can be time-consuming and often requires human expertise.

To address this gap, we explored two automatic methods to evaluate educational MCQs using the IWF rubric. The first method utilizes a rule-based approach to apply the rubric, making it easy to modify and maintain interpretability, while not requiring a large training dataset. Our second method relies on GPT-4, a large multimodal model capable of processing text inputs and producing text outputs, which has achieved human-level performance on various professional and academic benchmarks [30]. This second method prompts GPT-4 to apply the IWF criteria to the provided questions one at a time. Using student-generated questions from four distinct subject areas, we evaluated both methods and compared them to human expert evaluation that also utilized the IWF rubric. We investigate to what extent a rule-based multi-label classifier and GPT-4 can accurately identify IWFs in student-generated MCQs. With this setting, our research questions are as follows:

**RQ1**: How does automatic evaluation of educational MCQs using the IWF rubric compare to human evaluation?

**RQ2**: How does the performance of the automatic application of the IWF rubric vary across subject areas?

**RQ3**: Which IWFs are the most common in student-generated educational questions?

## 2 Related Work

### 2.1 Evaluating Assessment Questions with Student Data

Educational MCQs generated by instructors, students, or through automated methods are all susceptible to flaws that impact their efficacy and quality [33]. One challenge in evaluating MCQs' quality lies in determining what criteria are sufficient to indicate that a question has high quality and is effective for use in an educational context. To overcome this subjectivity, different item response theory and statistical methods have been utilized to evaluate student-generated MCQs, such as the Item Difficulty Index and the Discrimination Index [18, 21]. These techniques require post-hoc analysis of student performance data on the questions, as they use the percentage of students that answered the question correctly and measure the question's ability to differentiate between students who have a high level of knowledge and those who do not. However, this can be detrimental to the learning process, because if the questions being used have not been first vetted for their quality, then they may be poorly constructed which can negatively impact students' performance and achievement, causing these analyses to be inaccurate [6]. To avoid potential harm to student learning and save them time from answering questions with potentially low quality, questions should be evaluated prior to classroom usage. Towards this goal, recent research has investigated different methods for evaluating educational MCQs that utilize smaller amounts of student data, or no student data at all, to achieve a higher quality evaluation [7]. For instance, work by [39] utilized small amounts of student data along with their perceptions of certain questions to acquire a comprehensive evaluation of questions in an educational context.

## 2.2  Human Evaluation of Assessment Questions

The use of the IWF rubric is effective at evaluating educational questions, yet the application often requires substantial human effort and is time-consuming, especially when evaluating a large number of questions across multiple subject areas [16]. Human evaluation of both AQG and student-generated MCQs often serves as a benchmark for comparing how well a method classified the quality of the educational questions [4, 21]. This evaluation consists of one or more experts judging the questions based on "best practice" conventions, which can include their pure subjective judgment, such as indicating if they would use the question in their class, or the more formal application of a standardized rubric [11]. The most common human evaluation methods involve the application of a rubric. This practice helps to standardize the process, improve replicability, and decrease subjectivity [31]. While different rubrics have been employed for this process, the Item-Writing Flaws (IWFs) rubric containing 19 criteria for assessing educational questions has been standardized and evaluated via previous research [3, 28, 33]. One such study that utilized this 19-item rubric assessed the quality of over two thousand MCQs [37]. Utilizing two human evaluators, they determined that nearly half of the questions were deemed unacceptable for educational usage, due to having more than one IWF. In this case, the question difficulties may be skewed to be too easy or too hard, which in turn misleads students and related learning analytics [9].

## 2.3  Automatic Evaluation of Assessment Questions

The automatic evaluation of questions usually employs metrics such as readability and explainability, based on natural language processing (NLP) metrics like BLEU and METEOR [34]. However, recent research has shown that these metrics do not correlate with human evaluation and do not consider the pedagogical value of the questions [22]. To improve the classification accuracy of a question's quality, recent efforts have focused on training classification models using large datasets of student responses [29, 32]. These methods rarely utilize questions from complex domains that go beyond the cognitive process of recall [21]. Additionally, the model architectures used in these methods are often difficult to interpret because they have black-box evaluation criteria. For instance, a recent study achieved an 81.22% accuracy in classifying question quality using recurrent neural networks [32]. However, the classification criteria used a single label based on how "shallow" or "deep" human raters found the question to be, which may be subjective and hard to replicate. In another study, GPT-3 was used to assess the quality of short answer questions in post-secondary education and often overestimated the quality [27]. The authors proposed that this overestimation was due to the high complexity of the input questions, but verifying this conjecture was difficult due to the black-box model. To improve interpretability and accuracy, prior research has turned to rule-based classification methods, which can offer feedback on why a question was classified a certain way [40]. Rule-based methods break down the classification criteria into smaller components, making it easier to identify specific patterns and features within the data, and achieving similar or greater success than black-box models, such deep learning methods [41].

## 3 Methods

### 3.1 Dataset

The datasets used in this study were collected from a digital learning platform used by several public universities and community colleges in the western United States. The data comes from students using the platform in their respective courses during the 2020 and 2021 academic years. The four courses are introductory Chemistry, introductory Biochemistry, introductory Statistics, and a course on learning how to effectively collaborate, referred to as CollabU. Students in these courses were undergraduates, towards the beginning of their studies, and pursuing either a two- or four-year degree.

As students worked through the digital learning materials on the platform in their respective courses, they were prompted to create a multiple-choice question (MCQ). The prompt asked students to create a single MCQ about a topic they had recently learned in their course. Each MCQ consists of a question text, known as the *stem*, and four answer choices, one of which must be denoted as correct. The creation of this MCQ was done directly in the digital learning platform with no additional tools utilized. Students did not receive any assistance or feedback as they created their questions. Additionally, it was presented in the same visual manner as the other activities found on the platform. From each of these four courses, we randomly selected 50 student-generated questions to utilize for this study, resulting in a total of 200 MCQs.

### 3.2 Human Evaluation

In order to assess the quality of the student-generated MCQs, we utilized a series of guidelines for identifying Item-Writing Flaws (IWFs), which are based on a taxonomy of 31 multiple-choice item-writing guidelines [10]. The exact rubric we used for the study was a modified version that consists of 19 unique items and has been used and validated in previous studies [3, 8, 28]. Following [37], a question with 0 or 1 flaw identified by the rubric is considered *acceptable* and any questions with 2 or more flaws is considered *unacceptable*. This distinction is used to determine if a question could be utilized in a class as a formative assessment that the instructor would trust. A full description of the 19 items that make up the rubric can be found in Table 1.

Two item raters evaluated each student-generated MCQ, following the 19 IWF guidelines. Both raters had content-area expertise across all four domains, ample experience developing multiple-choice questions, and multiple prior training sessions in writing high quality assessments. Using the IWF rubric, the raters went through each of the 200 student-generated MCQs and applied the rubric to the question text and accompanying answer choices for each student contribution. The inter-rater reliability (IRR) values between the two evaluators for each rubric item are also reported in Table 1. It includes the percentage agreement and Cohen's Kappa κ statistic [26] as a measure of IRR for all rubric items. All items were at either a near perfect or substantial level of agreement between the raters. The two evaluators then met to resolve any disagreements in their evaluations and discussed discordant questions until they reached a consensus on the coding. We acknowledge that, despite the two expert evaluators' backgrounds and high IRR, they could still interpret the student-generated questions differently, based on their prior knowledge and linguistic preferences [2].

**Table 1.** The rubric of 19 Item-Writing Flaws used to evaluate the student-generated multiple-choice questions. The bracketed numbers indicate agreement percentage between raters and Cohen's κ value for each item.

| Item-Writing Flaw | Attributes of questions that do not contain the flaw |
|---|---|
| Ambiguous or unclear information (87.50%, κ = 0.66) | Questions and all options should be written in clear, unambiguous language |
| Implausible distractors (96.00%, κ = 0.82) | Make all distractors plausible as good items depend on having effective distractors |
| None of the above (100%, κ = 1.00) | Avoid none of the above as it only really measures students ability to detect incorrect answers |
| Longest option correct (97.00%, κ = 0.83) | Often the correct option is longer and includes more detailed information, which clues students to this option |
| Gratuitous information (89.50%, κ = 0.71) | Avoid unnecessary information in the stem that is not required to answer the question |
| True/false question (100%, κ = 1.00) | The options should not be a series of true/false statements |
| Convergence cues (89.50%, κ = 0.70) | Avoid convergence cues in options where there are different combinations of multiple components to the answer |
| Logical cues (88.00%, κ = 0.68) | Avoid clues in the stem and the correct option that can help the test-wise student to identify the correct option |
| All of the above (100%, κ = 1.00) | Avoid all of the above options as students can guess correct responses based on partial information |
| Fill-in-blank (100%, κ = 1.00) | Avoid omitting words in the middle of the stem that students must insert from the options provided |
| Absolute terms (100%, κ = 1.00) | Avoid the use of absolute terms (e.g. never, always, all) in the options as students are aware that they are almost always false |
| Word repeats (97.00%, κ = 0.83) | Avoid similarly worded stems and correct responses or words repeated in the stem and correct response |
| Unfocused stem (94.50%, κ = 0.79) | The stem should present a clear and focused question that can be understood and answered without looking at the options |
| Complex or K-type (94.00%, κ = 0.78) | Avoid questions that have a range of correct responses, that ask students to select from a number of possible combinations of the responses |
| Grammatical cues (92.50%, κ = 0.76) | All options should be grammatically consistent with the stem and should be parallel in style and form |
| Lost sequence (97.50%, κ = 0.89) | All options should be arranged in chronological or numerical order |
| Vague terms (98.50%, κ = 0.93) | Avoid the use of vague terms (e.g. frequently, occasionally) in the options as there is seldom agreement on their actual meaning |
| More than one correct (100%, κ = 1.00) | In single best-answer form, questions should have 1, and only 1, best answer |
| Negative worded (100%, κ = 1.00) | Negatively worded stems are less likely to measure important learning outcomes and can confuse students |

### 3.3 Rule-based Evaluation

The task of automatically applying the Item-Writing Flaws rubric to MCQs is a multi-label classification problem, as each question may be matched with several criteria [1]. In order to implement this automated method, we followed a rule-based approach that applies each individual rubric criteria via its own logic. Rule-based approaches have been used in similar educational tasks such as classifying the Bloom's Taxonomy of a question [12]. Such an approach is particularly effective when the problem suffers from a lack of training data, as is the case in the present study, due to a lack of public datasets containing questions that are evaluated for their educational quality [15]. Furthermore, a rule-based approach allows for vastly improved interpretability compared to traditional black-box classification approaches, such as neural networks [41].

Working alongside the human evaluators, we constructed a script that is composed of a programmatic method for each of the 19 IWF rubric criteria. It uses several Python libraries and three different pre-trained large language models (LLMs) to implement the 19 different criteria. The logic for many of the criteria involved string manipulation, such as checking if the *longest option was the correct answer*. Other criteria involved the use of standard NLP techniques, such as Named Entity Recognition or Part-of-Speech tagging [36] to help identify if a *word is repeated* between the question's stem and correct answer. The more challenging and advanced criteria, such as identifying if a question contains *implausible distractors*, involved the use of LLMs. For instance, a RoBERTa classifier pretrained on the Corpus of Linguistic Acceptability (CoLA) was utilized to help identify *ambiguous or unclear information* in a question's stem [20]. To determine if a question contained *more than one correct answer*, we leveraged GPT-4's capabilities for question answering [30]. For a more detailed explanation of the programmatically implemented 19 IWFs criteria, the final code is made publicly available[1], however the student question data is currently private and can be made available upon request.

### 3.4 GPT-4 Evaluation

The second automatic evaluation method utilizes GPT-4, a transformer-based multimodal model pre-trained to predict the next token in a document [30]. We utilize

---
[1] https://github.com/StevenJamesMoore/ECTEL23/blob/main/IWF.ipynb

GPT-4 as our second automated method as it has achieved human-level performance on academic tasks, such as standardized college-level exams in Psychology, History, and Math. It has also achieved state-of-the-art performance on traditional machine learning benchmarks, such as the MMLU, which consists of 57 tasks from a variety of domains that are used to demonstrate a model's extensive world knowledge and problem-solving ability [13]. What also makes GPT-4 unique compared to many other language models, is the ability to follow natural language prompts to perform specific tasks [24]. These prompts serve as instructions for the model to perform, such as providing the model with a rubric and multiple-choice question and then prompting it to apply the rubric criteria to the question.

The exact wording of the prompts provided to GPT-4 can drastically influence the output the model provides [23]. Towards this end, our task involved providing GPT-4 with a single IWF rubric criteria at a time and having it state if the provided question satisfied that given criteria or not. The names of the criteria we used as well as their definitions are nearly identical to the ones shown in Table 1. We opted to directly use the prompts rather than continually engineering prompts to determine the best output. While we believe refinement of the prompts is valuable future work, we wanted to see how well GPT-4 would perform applying the same IWF rubric and giving it instructions akin to what would be provided to a human evaluator. Specifically, the prompt we provided GPT-4 for each IWF rubric criteria and question states: *Begin your response with yes or no, does this multiple-choice question satisfy the criteria relating to {criteria}: {definition}? Explain why. {question}*. The rubric criteria, definition of the criteria, and the multiple-choice question including all answer options are input into the prompt respectively. Additionally, we utilized the default parameters of the model and accessed it using the GPT-4 API via the Python programming language.

Note, the prompt instructions also asked GPT-4 to provide an explanation as to why a question satisfies or violates the criteria, which was done to encourage a more thorough and accurate response from the model [17]. A human evaluator went through each of the responses and coded them as GPT-4 indicating if the criteria was satisfied or violated. Although we had originally intended to use a simple "Yes" or "No" response to indicate whether the criteria were met, we found that this approach was not always clear in distinguishing whether the criteria had been violated or satisfied.

## 4 Results

### 4.1 Automatic Methods versus Human Identification

The 19 IWF criteria were automatically applied to all 200 student-generated questions, resulting in a total of 3800 classifications. The rule-based method matched 90.87% of human classifications, achieving an exact match ratio of 15%, where all of the 19 IWF criteria matched the human evaluation for the question. The GPT-4 method matched 78.89% of human classifications, achieving an exact match ratio of 12%. We also considered the Hamming loss, which is a measure of the difference between two sets of binary labels and calculated as the fraction of labels that are incorrectly predicted [38]. The rule-based method achieved a Hamming Loss of 0.09 and the GPT-4 method achieved a Hamming Loss of 0.21, indicating that on average 9% and 21% of the flaws were misclassified respectively. Table 2 displays the number of IWFs assigned to questions for each evaluation method grouped by counts. A paired t-test showed a small significant difference in the number of IWFs identified for each question by the human (*M=1.6, SD=1.3*) and rule-based (*M=2.1, SD=1.4*) methods, *t(199)=5.59, p<.001*. The rule-based evaluation method more commonly identified potential flaws in the questions compared to the humans. Another paired t-test showed a significant difference in the number of IWFs identified for each question by the human (*M=1.6, SD=1.3*) and GPT-4 (*M=4.2, SD=3.4*) methods, *t(199)=11.8, p<.001*. The GPT-4 evaluation identified even more potential flaws in the questions compared to both the human and rule-based methods. While the human and rule-based methods never found more than six IWFs per question, the GPT-4 method found up to thirteen.

**Table 2.** Counts of IWFs per question from all three evaluation methods.

| Number of Flaws | 0 | 1 | 2 | 3 | 4 | 5 | 6 | 7 | 8 | 9 | 10 | 11 | 12 | 13 |
|---|---|---|---|---|---|---|---|---|---|---|---|---|---|---|
| Human Evaluation | 39 | 72 | 42 | 28 | 9 | 6 | 2 | 0 | 0 | 0 | 0 | 0 | 0 | 0 |
| Rule-based Evaluation | 23 | 49 | 56 | 34 | 27 | 8 | 1 | 0 | 0 | 0 | 0 | 0 | 0 | 0 |
| GPT-4 Evaluation | 30 | 23 | 25 | 24 | 13 | 17 | 7 | 19 | 11 | 11 | 9 | 7 | 1 | 1 |

When the quality of the questions were labeled as acceptable (< 2 IWFs) or unacceptable (≥ 2 IWFs), a chi-square test revealed there was a significant relationship between the question quality and three evaluation methods, $\chi^2(2,N=200)=36.64, p<.001$. Between the three methods, GPT-4 was more likely to evaluate a question as having unacceptable quality. The human evaluation identified 111 acceptable and 89 unacceptable questions, while the rule-based evaluation matched 130 (65%) of these (57 acceptable, 73 unacceptable). The GPT-4 evaluation matched 123 (62%) of the human quality evaluations (44 acceptable, 79 unacceptable). Figure 1 shows the confusion matrices for the quality classifications based on the number of IWFs found in each question between the human and rule-based evaluation and the human and GPT-4 evaluation.

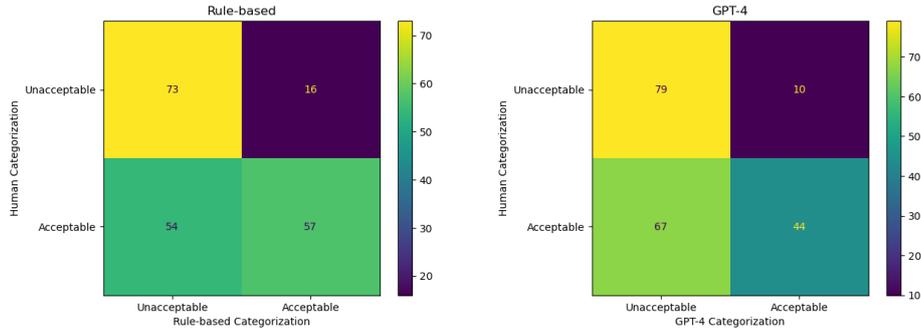

**Fig. 1.** Confusion matrices for the classification of a question's quality for the rule-based method (left) and the GPT-4 method (right).

### 4.2 Impact of the Domain

The automatic evaluation methods, rule-based and GPT-4, performed differently across criteria and domains, with the rule-based method outperforming GPT-4 on all four domains. Table 3 shows the performance of all three evaluation methods across all four domains. Between the datasets, we use F1 scores to evaluate success. Since a majority of the questions meet the criteria rather than violate them, the F1 score provides a better measure over accuracy, as it includes false negatives and false positives. From Table 3, we observe that the rule-based and GPT-4 methods commonly matched human evaluation for some criteria, such as *none of the above* and *negative worded*, and performed poorly for other criteria, such as *logical cues* and *more than one correct*. In particular, the rule-based method achieves high F1 scores for *longest option correct* and *true/false question* compared to GPT-4. Note, the rules for these two criteria can be easily implemented programmatically, as they only check for text length and keywords.

Both the rule-based and GPT-4 methods have a lower micro-average F1 score, the computed proportion of correctly classified observations out of all observations, for the Chemistry and Biochemistry courses compared to Statistics and CollabU. This may be in part due to the similar domains of these science courses, where the human evaluators focused more on the objective of the questions, rather than the grammar, while the automated methods did not. Additionally, some of the poor performance related to F1 scores is due to the small number of that flaw being found in the questions. For instance, *gratuitous information* and *vague terms* have poor performance by F1 score, but those flaws are quite rare across all four courses. Ultimately, the rule-based method outperformed GPT-4, by measure of micro-average F1 score, across all four domains.

**Table 3.** The count of flaws (N) and performance (F1) of the human evaluation (Hum) compared to both the rule-based (Rule) and GPT-4 (GPT) methods across all four domains. A dash (-) in the table indicates that the flaw was not present in any question of that dataset based on human evaluation and therefore no F1 score is computed.

| Item-Writing Flaws | | Chemistry | | | Biochemistry | | | Statistics | | | CollabU | | |
|---|---|---|---|---|---|---|---|---|---|---|---|---|---|
| | | Hum | Rule | GPT | Hum | Rule | GPT | Hum | Rule | GPT | Hum | Rule | GPT |
| ambiguous information | N | 12 | 2 | 11 | 7 | 4 | 25 | 14 | 11 | 25 | 7 | 7 | 40 |
| | F1 | - | 0.14 | 0.61 | - | 0.00 | 0.25 | - | 0.40 | 0.41 | - | 0.29 | 0.30 |
| implausible distractors | N | 10 | 6 | 17 | 3 | 9 | 16 | 8 | 9 | 17 | 25 | 21 | 29 |
| | F1 | - | 0.12 | 0.30 | - | 0.33 | 0.11 | - | 0.82 | 0.16 | - | 0.83 | 0.78 |
| none of the above | N | 7 | 8 | 6 | 3 | 3 | 3 | 1 | 1 | 1 | 1 | 1 | 1 |
| | F1 | - | 0.80 | 0.62 | - | 1.00 | 0.67 | - | 1.00 | 0.00 | - | 1.00 | 1.00 |
| longest option correct | N | 5 | 5 | 6 | 2 | 2 | 11 | 3 | 3 | 6 | 9 | 10 | 13 |
| | F1 | - | 0.80 | 0.18 | - | 1.00 | 0.00 | - | 1.00 | 0.44 | - | 0.95 | 0.36 |
| gratuitous information | N | 0 | 0 | 2 | 7 | 5 | 25 | 3 | 0 | 18 | 0 | 0 | 19 |
| | F1 | - | - | 0.00 | - | 0.67 | 0.38 | - | 0.00 | 0.29 | - | - | 0.00 |
| true/false question | N | 2 | 3 | 1 | 6 | 11 | 11 | 1 | 1 | 3 | 3 | 3 | 4 |
| | F1 | - | 0.80 | 0.67 | - | 0.59 | 0.24 | - | 1.00 | 0.00 | - | 1.00 | 0.29 |
| convergence cues | N | 2 | 12 | 12 | 12 | 36 | 13 | 9 | 11 | 18 | 10 | 16 | 16 |
| | F1 | - | 0.14 | 0.14 | - | 0.50 | 0.24 | - | 0.80 | 0.30 | - | 0.77 | 0.23 |
| logical cues | N | 2 | 2 | 9 | 2 | 6 | 15 | 3 | 1 | 18 | 10 | 0 | 23 |
| | F1 | - | 0.00 | 0.00 | - | 0.00 | 0.00 | - | 0.00 | 0.10 | - | 0.00 | 0.36 |
| all of the above | N | 2 | 0 | 1 | 3 | 0 | 2 | 1 | 0 | 1 | 2 | 1 | 3 |
| | F1 | - | 0.00 | 0.67 | - | 0.00 | 0.80 | - | 0.00 | 1.00 | - | 0.67 | 0.80 |
| fill-in-the-blank | N | 2 | 2 | 3 | 4 | 3 | 2 | 0 | 0 | 0 | 1 | 1 | 4 |
| | F1 | - | 1.00 | 0.00 | - | 0.86 | 0.67 | - | - | - | - | 1.00 | 0.40 |
| | N | 2 | 6 | 1 | 6 | 19 | 9 | 0 | 2 | 1 | 7 | 10 | 6 |
| absolute terms | F1 | - | 0.00 | 0.00 | - | 0.40 | 0.27 | - | 0.00 | 0.00 | - | 0.71 | 0.62 |
| word repeats | N | 0 | 0 | 2 | 8 | 0 | 7 | 0 | 0 | 6 | 3 | 2 | 5 |
| | F1 | - | - | 0.00 | - | 0.00 | 0.13 | - | - | 0.00 | - | 0.00 | 0.25 |
| unfocused stem | N | 0 | 2 | 7 | 5 | 4 | 15 | 7 | 3 | 17 | 4 | 4 | 28 |
| | F1 | - | 0.00 | 0.00 | - | 0.44 | 0.30 | - | 0.40 | 0.50 | - | 0.75 | 0.25 |
| complex or K-type | N | 2 | 0 | 6 | 6 | 9 | 6 | 3 | 2 | 14 | 1 | 1 | 27 |
| | f1 | - | 0.00 | 0.25 | - | 0.53 | 0.50 | - | 0.80 | 0.24 | - | 1.00 | 0.07 |
| grammatical cues | N | 1 | 3 | 17 | 13 | 24 | 14 | 5 | 13 | 20 | 11 | 18 | 36 |
| | F1 | - | 0.00 | 0.11 | - | 0.65 | 0.30 | - | 0.44 | 0.32 | - | 0.76 | 0.38 |
| lost sequence | N | 0 | 2 | 12 | 2 | 0 | 12 | 11 | 11 | 17 | 0 | 0 | 7 |
| | F1 | - | 0.00 | 0.00 | - | 0.00 | 0.29 | - | 0.91 | 0.43 | - | - | 0.00 |
| vague terms | N | 0 | 0 | 0 | 2 | 0 | 4 | 0 | 0 | 3 | 0 | 0 | 10 |
| | F1 | - | - | - | - | 0.00 | 0.00 | - | - | 0.00 | - | - | 0.00 |
| more than one correct | N | 0 | 13 | 5 | 0 | 4 | 14 | 0 | 20 | 16 | 4 | 10 | 22 |
| | F1 | - | 0.00 | 0.00 | - | 0.00 | 0.00 | - | 0.00 | 0.00 | - | 0.43 | 0.15 |
| negative worded | N | 0 | 0 | 0 | 8 | 14 | 15 | 2 | 3 | 2 | 6 | 7 | 6 |
| | F1 | - | - | - | - | 0.64 | 0.70 | - | 0.80 | 1.00 | - | 0.92 | 1.00 |
| micro-avg | | - | 0.30 | 0.25 | - | 0.48 | 0.28 | - | 0.56 | 0.30 | - | 0.70 | 0.36 |
| totals | | 49 | 66 | 118 | 99 | 149 | 219 | 71 | 91 | 203 | 104 | 112 | 299 |

### 4.3 Common Item-Writing Flaws

The most frequently identified violated criteria varied across the three methods, although in some domains the rule-based and GPT-4 methods had similar classifications to the human evaluation. Table 3 shows that the *implausible distractor* criteria was violated the most across all questions in human evaluation, whereas *vague terms* was the least violated. On the other hand, the rule-based method found *convergence cues* to be the most commonly violated criteria and *vague terms* to also be the least violated. As for the GPT-4 method, the most commonly violated criteria was *ambiguous unclear information*, and *all of the above* was the least violated. Although the most frequently violated criteria varied across all three methods, the rule-based and GPT-4 methods shared similar results with human evaluation. Specifically, the rule-based method's most violated criteria ranked as the third most violated criteria in human evaluation, while the GPT-4 method's most violated criteria ranked second.

A Pearson correlation coefficient showed there was a significant positive correlation between the number of flaws identified for each criteria between the rule-based and human evaluations for Biochemistry (*r(17) =.747, p<.001*), Statistics (*r(17)=.496, p<.05*), and CollabU (*r(17)=.833, p<.001*). However, this correlation was not found to be significant for Chemistry (*r(17)=.191, p=.433*). A Pearson correlation coefficient was also computed for the number of flaws identified for each criteria between the GPT-4 and human evaluations. There was a significant positive correlation found for Statistics (*r(17)=.756, p<.001*), and CollabU (*r(17)=.4702, p<.05*). No significant correlation was found for Chemistry (*r(17) =.443, p=.057*) and Biochemistry (*r(17) =.262, p=.278*). This suggests that for Statistics and CollabU, both automated methods identified similar trends in the violated criteria – i.e., if a flaw was commonly found by human evaluation, it was also likely to be commonly found by the automated methods.

# 5    Discussion

In this work, we developed an automatic rule-based method and assessed its performance compared to GPT-4 and human annotation for evaluating the quality of educational MCQs using the IWF rubric. In contrast to prior research, we employed criteria that pertain directly to the pedagogical value of the question across multiple dimensions. We found that this method can perform at a level comparable to human evaluation for certain rubric criteria and outperforms GPT-4 in the same task across all rubric criteria. The rule-based method was effective in evaluating questions across four distinct subject areas, even with the presence of domain-specific jargon. When comparing the results of our automatic evaluation methods to human evaluation, we identified commonly found IWFs in student-generated questions across the four subject areas. Our results suggest that using a rule-based multi-label classification method can achieve a high level of accuracy while also maintaining interpretability, which the GPT-4 method lacks.

Both of the automatic methods' classifications were stricter, in the sense that they assigned many more IWFs to the student-generated questions than human experts, particularly GPT-4. However, this is preferable to being less strict, as guaranteeing high-quality questions during the evaluation process is crucial so as to not disrupt student learning. Additionally, both of the automatic methods could easily help filter out questions whose quality is too low for human review, e.g., if a question has four or more IWFs, it would likely take substantial time to review and could be dropped. This filtering capability of the rule-based method is supported by our results showing that it matches 65% of human classification when categorizing questions as *acceptable* or *unacceptable*, based on the IWF count. This method of binary classification of quality is commonly used in MCQ evaluation and has a performance level comparable to other models using similar educational datasets [29, 32]. Additionally, these automatic methods could be identifying criteria that were missed by the human evaluators, rather than misclassifying questions with the IWF rubric criteria.

While GPT-4's training data included material from the four course domains used in this study, its black-box nature poses challenges in interpreting why it might be misclassifying specific IWF criteria [30]. For instance, the GPT-4 method achieves extremely low F1 scores for *gratuitous information*, *unfocused stem*, and *vague terms*, all of which relate to the question's stem being unnecessarily verbose. Our analysis revealed that GPT-4 identifies a significantly high number of these three flaws across questions in each domain compared to the human and rule-based methods. This could mean that GPT-4 is mistakenly combining these criteria due to their similarity, marking them all as violated based on a single flaw. In contrast, the rule-based method can be designed to implement each criteria explicitly without overlapping with other flaws.

Across the four different subject areas utilized in this study, we found that both the rule-based and GPT-4 methods performed better on Statistics and CollabU, compared to Chemistry and Biochemistry. This may be in part due to the latter two domains containing questions that use more terminologies and jargon, making some of the NLP techniques less effective [7]. Interestingly, the rule-based method achieved more than double the micro-average F1 score of the GPT-4 method in CollabU. GPT-4's worse performance in this case may be due to proper nouns being included in the question text. The human evaluators familiar with the course content would find the usage of the proper nouns acceptable and the rule-based method does not leverage proper nouns in many of the criteria. However, GPT-4 may identify these as errors in the question as it lacks the necessary context to know if they are essential to the question or not.

Compared to the other course domains, CollabU, a course on learning how to effectively collaborate, may have more recall-level student-generated questions. In contrast, the other domains may include more complex questions that involve formulas or numbers that are challenging to decipher for the automatic methods. Criteria such as *lost sequence* are also more applicable to domains such as Chemistry or Statistics as they may include question options that are purely numerical, causing the arrangement of options to matter. Additionally, both automatic methods performed the worst in Chemistry and Biochemistry, two closely related science courses. IWF criteria such as *grammatical cues* and *convergence cues* were excessively identified by both methods compared to the human evaluators. With the subjectivity that arises from human evaluation, even when applying a standardized rubric, it is possible the evaluators were less focused on grammar in this domain and more focused on the objective of the question, prioritizing *what* was being asked more than *how* it was being asked. This highlights the need for automatic evaluation methods that can focus on both the syntax and the question's content that is critical to the domain and pedagogy. In contrast to human evaluation, automatic methods scale easier, reduce human subjectivity leading to enhanced replicability, and can be used by individuals without domain expertise.

Finally, the rule-based method demonstrated effectiveness in identifying IWFs that are common in accordance with human evaluation in each of the datasets analyzed. In line with previous research, *ambiguous or unclear information* and *implausible distractors* were two of the most identified flaws across all questions by the human, rule-based, and GPT-4 methods [33, 37]. Our analysis revealed that 50% of the questions in the CollabU dataset exhibited the *implausible distractors* flaw. Again, this may be attributed to the recall-based nature of the material in this domain, which could make it challenging for students to generate plausible alternative options for the questions. In contrast to [33, 37], our datasets contained a high percentage of questions with the *convergence cues* flaw. This might be a result of the digital interface that students used to construct MCQs in our study, as it might have encouraged them to copy the correct answer and then modify it, leading to the prevalence of this flaw. In turn, these findings can inform teachers of the common flaws that they should focus on when refining student-generated MCQs and providing them with feedback on the task.

We expected both automatic methods to perform highly on the IWF criteria *more than one correct*, as they both leverage GPT-4, which has achieved success in these course domains [30]. However, the presence of different flaws, such as incorrect grammar or the inclusion of proper nouns, may cause the question to be confusing, potentially misleading GPT-4 into incorrectly answering some of the questions. Additionally, while criteria such as *none of the above* and *fill-in-the-blank* might initially appear to be easy to achieve near perfect accuracy on, they can give both the rule-based and GPT-4 methods difficulty. For instance, the rule-based method, which uses keyword matching, might not properly detect *none of the above* if there is a spelling mistake or if there are extra words amongst the given option. Similarly, GPT-4 was often overzealous at detecting these flaws, as at times it interpreted different answer options as effectively containing text akin to *none of the above*, despite it not explicitly being an option.

13## 6 Limitations and Future Work

We identified several limitations to our study that may be addressed in future research. First, our study relies on several datasets of student-generated questions, whose quality may vary by the subject area and individual students. Analyzing educational MCQs from other domains that contain a different variety of flaws could lead to more holistic and generalizable findings. It should be noted that the classification of questions in our study is inherently subjective due to the nature of human evaluation. To mitigate this, we employed a standardized IWF rubric and achieved a high inter-rater reliability (IRR) for each criteria. However, it is possible that different evaluators may arrive at different results. The code implementation used to identify the item-writing flaws could be adjusted to achieve different results. For example, variations in threshold for cosine similarity, utilizing an alternative implementation of a method from a different library, or rewording the GPT-4 prompts could affect the outcome.

Finally, the use of GPT-4 poses challenges with replicability, despite providing the prompts and default hyperparameters used in this research. One challenge is that the output of GPT-4 still requires human evaluation to interpret and verify what the model intended, as even when it is prompted for specific phrasing, it may still respond in a conflating manner. Another related challenge is that the model is both inherently random to some degree and still under development, meaning at a future point in time it might perform differently given the same tasks as this research. In order to promote transparency and reproducibility of our research, we have open-sourced our code. This allows for full visibility into the logic used for classifying each item-writing flaw and maintains interpretability so that other researchers can easily make any desired modifications. A promising future direction of this work is to both improve the classification accuracy of these flaws and extend the automatic methods to also provide suggestions for addressing the identified flaws. While GPT-4 may have not been as accurate as the rule-based method for identifying the flaws, it can provide explanations and suggest improvements to the questions.

## 7 Conclusion

In this paper, we proposed a novel rule-based method for automatically evaluating the quality of educational multiple-choice questions using criteria from the Item-Writing Flaws rubric. The results indicate that the rule-based method accurately assesses the quality of student-generated questions across multiple distinct subject areas and highlights the occurrence of different flaws in questions across these domains. It outperforms GPT-4 in applying the Item-Writing Flaws rubric across all four domains when compared to human evaluation. Both automated methods demonstrate how certain flaws may be easier or harder to identify, depending on the subject area. We contribute a categorization and comparison of item-writing flaws found in student-generated questions across four different subject areas. These results provide a valuable baseline performance measure for future research. This work also opens further opportunities for developing open and interpretable methods for evaluating educational questions by pedagogical values.

# References

1. Alazaidah R, Thabtah F, Al-Radaideh Q (2015) A multi-label classification approach based on correlations among labels. Int J Adv Comput Sci Appl 6:52–59
2. Amidei J, Piwek P, Willis A (2018) Rethinking the Agreement in Human Evaluation Tasks. In: Proceedings of the 27th International Conference on Computational Linguistics.
3. Breakall J, Randles C, Tasker R (2019) Development and use of a multiple-choice item writing flaws evaluation instrument in the context of general chemistry. Chem Educ Res Pract 20:369–382
4. Brown GT, Abdulnabi HH (2017) Evaluating the quality of higher education instructor-constructed multiple-choice tests: Impact on student grades. In: Frontiers in Education. Frontiers Media SA, p 24
5. Butler AC (2018) Multiple-choice testing in education: Are the best practices for assessment also good for learning? J Appl Res Mem Cogn 7:323–331
6. Clifton SL, Schriner CL (2010) Assessing the quality of multiple-choice test items. Nurse Educ 35:12–16
7. Cochran K, Cohn C, Hutchins N, Biswas G, Hastings P (2022) Improving automated evaluation of formative assessments with text data augmentation. In: International Conference on Artificial Intelligence in Education. Springer, pp 390–401
8. Danh T, Desiderio T, Herrmann V, Lyons HM, Patrick F, Wantuch GA, Dell KA (2020) Evaluating the quality of multiple-choice questions in a NAPLEX preparation book. Curr Pharm Teach Learn
9. Downing SM (2005) The effects of violating standard item writing principles on tests and students: the consequences of using flawed test items on achievement examinations in medical education. Adv Health Sci Educ 10:133–143
10. Haladyna TM (2004) Developing and Validating Multiple-choice Test Items. Psychology Press
11. Haladyna TM, Downing SM, Rodriguez MC (2002) A review of multiple-choice item-writing guidelines for classroom assessment. Appl Meas Educ 15:309–333
12. Haris SS, Omar N (2012) A rule-based approach in Bloom's Taxonomy question classification through natural language processing. In: 2012 7th international conference on computing and convergence technology (ICCCT). IEEE, pp 410–414
13. Hendrycks D, Burns C, Basart S, Zou A, Mazeika M, Song D, Steinhardt J Measuring Massive Multitask Language Understanding. In: International Conference on Learning
14. Horbach A, Aldabe I, Bexte M, de Lacalle OL, Maritxalar M (2020) Linguistic appropriateness and pedagogic usefulness of reading comprehension questions. In: Proceedings of The 12th Language Resources and Evaluation Conference. pp 1753–1762
15. Hüllermeier E, Fürnkranz J, Loza Mencia E, Nguyen V-L, Rapp M (2020) Rule-based multi-label classification: Challenges and opportunities. In: International Joint Conference on Rules and Reasoning. Springer, pp 3–19
16. Ji T, Lyu C, Jones G, Zhou L, Graham Y (2022) QAScore—An Unsupervised Unreferenced Metric for the Question Generation Evaluation. Entropy 24:1514
17. Kasneci E, Seßler K, Küchemann S, Bannert M, Dementieva D, Fischer F, Gasser U, Groh G, Günnemann S, Hüllermeier E, others (2023) ChatGPT for good? On opportunities and challenges of large language models for education. Learn Individ Differ 103:102274
18. Khairani AZ, Shamsuddin H (2016) Assessing Item Difficulty and Discrimination Indices of Teacher-Developed Multiple-Choice Tests. In: Assessment for Learning Within and Beyond the Classroom. Springer, pp 417–426
19. Khosravi H, Demartini G, Sadiq S, Gasevic D (2021) Charting the design and analytics agenda of learnersourcing systems. In: LAK21: 11th International Learning Analytics and Knowledge Conference. pp 32–42
20. Krishna K, Wieting J, Iyyer M (2020) Reformulating Unsupervised Style Transfer as

Paraphrase Generation. In: Proceedings of the 2020 Conference on Empirical Methods in Natural Language Processing (EMNLP). pp 737–762
21. Kurdi G, Leo J, Parsia B, Sattler U, Al-Emari S (2020) A systematic review of automatic question generation for educational purposes. Int J Artif Intell Educ 30:121–204
22. van der Lee C, Gatt A, van Miltenburg E, Krahmer E (2021) Human evaluation of automatically generated text. Comput Speech Lang 67:101151
23. Lee P, Bubeck S, Petro J (2023) Benefits, Limits, and Risks of GPT-4 as an AI Chatbot for Medicine. N Engl J Med 388:1233–1239
24. Liu Y, Iter D, Xu Y, Wang S, Xu R, Zhu C (2023) GPTEval: NLG Evaluation using GPT-4 with Better Human Alignment. ArXiv Prepr ArXiv230316634
25. Lu OH, Huang AY, Tsai DC, Yang SJ (2021) Expert-Authored and Machine-Generated Short-Answer Questions for Assessing Students Learning Performance. Educ Technol Soc
26. McHugh ML (2012) Interrater reliability: the kappa statistic. Biochem Medica 22:276–282
27. Moore S, Nguyen HA, Bier N, Domadia T, Stamper J (2022) Assessing the Quality of Student-Generated Short Answer Questions Using GPT-3. 17th European Conference on Technology Enhanced Learning, EC-TEL 2022, Toulouse, France, September 12–16, 2022, Proceedings. Springer, pp 243–257
28. Moore S, Nguyen HA, Stamper J (2021) Examining the Effects of Student Participation and Performance on the Quality of Learnersourcing Multiple-Choice Questions. In: Proceedings of the Eighth ACM Conference on Learning@ Scale. pp 209–220
29. Ni L, Bao Q, Li X, Qi Q, Denny P, Warren J, Witbrock M, Liu J (2022) Deepqr: Neural-based quality ratings for learnersourced multiple-choice questions. In: Proceedings of the AAAI Conference on Artificial Intelligence. pp 12826–12834
30. OpenAI: GPT-4 Technical Report (2023), http://arxiv.org/abs/2303.08774.
31. Pugh D, De Champlain A, Gierl M, Lai H, Touchie C (2020) Can automated item generation be used to develop high quality MCQs that assess application of knowledge? Res Pract Technol Enhanc Learn 15:1–13
32. Ruseti S, Dascalu M, Johnson AM, Balyan R, Kopp KJ, McNamara DS, Crossley SA, Trausan-Matu S (2018) Predicting question quality using recurrent neural networks. In: International conference on artificial intelligence in education. Springer, pp 491–502
33. Rush BR, Rankin DC, White BJ (2016) The impact of item-writing flaws and item complexity on examination item difficulty and discrimination value. BMC Med Educ 1-10
34. Scialom T, Staiano J (2020) Ask to Learn: A Study on Curiosity-driven Question Generation. In: Proceedings of the 28th International Conference on Computational Linguistics. pp 2224–2235
35. Singh A, Brooks C, Doroudi S (2022) Learnersourcing in Theory and Practice: Synthesizing the Literature and Charting the Future. In: Proceedings of the Ninth ACM Conference on Learning@ Scale. pp 234–245
36. Straková J, Straka M, Hajic J (2014) Open-source tools for morphology, lemmatization, POS tagging and named entity recognition. In: Proceedings of 52nd Annual Meeting of the Association for Computational Linguistics: System Demonstrations. pp 13–18
37. Tarrant M, Knierim A, Hayes SK, Ware J (2006) The frequency of item writing flaws in multiple-choice questions used in high stakes nursing assessments. Nurse Educ Today
38. Tsoumakas G, Katakis I (2007) Multi-label classification: An overview. Int J Data Warehous Min IJDWM 3:1–13
39. Van Campenhout R, Hubertz M, Johnson BG (2022) Evaluating AI-Generated Questions: A Mixed-Methods Analysis Using Question Data and Student Perceptions. In: International Conference on Artificial Intelligence in Education. Springer, pp 344–353
40. van der Waa J, Nieuwburg E, Cremers A, Neerincx M (2021) Evaluating XAI: A comparison of rule-based and example-based explanations. Artif Intell 291:103404
41. Wang Z, Zhang W, Liu N, Wang J (2021) Scalable rule-based representation learning for interpretable classification. Adv Neural Inf Process Syst 34:30479–30491